\documentclass[preprint,12pt,authoryear]{elsarticle}

\usepackage[utf8]{inputenc}
\usepackage{multirow}
\usepackage{url}
\usepackage{booktabs}
\usepackage{amsmath}
\usepackage{latexsym}
\usepackage{color}
\usepackage{amsfonts}
\usepackage{graphicx}
\usepackage{todonotes}
\usepackage{scrextend}
\usepackage{caption}
\usepackage{subcaption}
\usepackage{comment}
\usepackage{microtype}

\renewcommand{\cite}[1]{\citep{#1}}

\newcommand{\alert}[1]{{\color{black}{#1}}}
\newcommand{\alertzwei}[1]{{\color{black}{#1}}}
\newcommand{\ex}[1]{\textit{``#1''}}

\journal{Preprint}

\begin{document}

\begin{frontmatter}

\title{A Resource-Light Method for Cross-Lingual \\ Semantic Textual Similarity}

\cortext[cor1]{Corresponding author}

\author[uma]{Goran Glava\v{s}\corref{cor1}}
\ead{goran@informatik.uni-mannheim.de}

\author[sym,uv]{Marc Franco-Salvador}
\ead{mfranco@prhlt.upv.es}

\author[uma]{Simone P. Ponzetto}
\ead{simone@informatik.uni-mannheim.de}

\author[uv]{Paolo Rosso}
\ead{prosso@dsic.upv.es} 

\address[uma]{Data and Web Science Group, School of Business Informatics and Matemathics, University of Mannheim, B6 26, DE-68159 Mannheim, Germany}

\address[sym]{Symanto Research, Pretzfelder Strasse 15, DE-90425, Nuremberg, Germany}

\address[uv]{Pattern Recognition and Human Language Technology Research Center, Universitat Politècnica de València, Camino de Vera s/n, ES-46022, Valencia, Spain}

\begin{abstract}

Recognizing semantically similar sentences or paragraphs across languages is beneficial for many tasks, ranging from cross-lingual information retrieval and plagiarism detection to machine translation. Recently proposed methods for predicting cross-lingual semantic similarity of short texts, however, make use of tools and resources (e.g., machine translation systems, syntactic parsers or named entity recognition) that for many languages (or language pairs) do not exist.
In contrast, we propose an unsupervised and a very resource-light approach for measuring semantic similarity between texts in different languages. To operate in the bilingual (or multilingual) space, we project continuous word vectors (i.e., word embeddings) from one language to the vector space of the other language via the linear translation model. We then align words according to the similarity of their vectors in the bilingual embedding space and investigate different unsupervised measures of semantic similarity exploiting bilingual embeddings and word alignments. 
%
%
Requiring only a limited-size set of word translation pairs between the languages, the proposed approach is applicable to virtually any pair of languages for which there exists a sufficiently large corpus, required to learn monolingual word embeddings.
%
Experimental results on three different datasets for measuring semantic textual similarity show that our simple resource-light approach reaches performance close to that of supervised and resource-intensive methods, displaying stability across different language pairs. Furthermore, we evaluate the proposed method on two extrinsic tasks, namely extraction of parallel sentences from comparable corpora and cross-lingual plagiarism detection, and show that it yields performance comparable to those of complex resource-intensive state-of-the-art models for the respective tasks.  

\end{abstract}


\end{frontmatter}

\section{Introduction}

There are many applications in natural language processing (NLP) -- machine translation \cite{resnik2003web,aziz2011fully}, cross-lingual information retrieval \cite{francosalvador-rosso-navigli:2014:EACL,DBLP:conf/sigir/VulicM15}, or plagiarism detection \cite{potthast2011cross,FrancoSalvador2016:KNOSYS}, to name a few -- that could directly exploit methods for measuring semantic similarity between short texts in different languages. 
Although recent years have seen a great amount of work on measuring semantic textual similarity (STS) of short texts (i.e., the task of determining the degree of semantic equivalence between short texts), the vast majority of these efforts focused on monolingual STS. Interest for STS has to the largest extent been instigated by the series of dedicated Sem\-E\-val tasks \citep[\textit{inter alia}]{agirre2012semeval,agirrea2015semeval}. 

\alert{The task of measuring semantic textual similarity (STS) amounts to estimating the degree to which two short texts are semantically related or associated, ranging from semantic equivalence (i.e., the meaning of the two short texts is exactly the same) to complete unrelatedness (i.e., the meaning of one short text is completely unrelated to the meaning of the other).} 
\alert{As such, the STS task is defined differently from two closely-related tasks: (1) recognizing textual entailment (RTE, used interchangeably with natural language inference) \cite{dagan2010recognizing,bowman-EtAl:2015:EMNLP} and paraphrase detection (PD) \cite{socher2011dynamic,madnani2012re}. In RTE, automatic methods need to recognize one-directional textual entailment between a pair of sentences (e.g., \ex{cat is purring} entails \ex{animal is making a sound}, but not vice-versa). PD, on the other hand, is the binary classification task in which classifiers need to determine whether two texts are paraphrases, i.e., semantically equivalent statements, or not.     
The STS aims to measure the degree of overlap in meaning between the two texts  (e.g., \ex{cat is purring} is arguably semantically more related to \ex{dog is barking} than to \ex{turtle is running}).}

Surprisingly -- unlike for RTE, for which several Cross-Lingual (CL) methods have been proposed \cite{castillo2011wordnet,mehdad2011using,negri2012semeval} -- the first approaches for cross-lingual STS have only been proposed most recently \cite{agirre-EtAl:2016:SemEval1,brychcin-svoboda:2016:SemEval,jimenez:2016:SemEval}. This is despite the rather obvious applicability of cross-lingual STS in extracting parallel sentences for machine translation (MT) \cite{resnik2003web,aziz2011fully}, cross-lingual information retrieval \cite{francosalvador-rosso-navigli:2014:EACL,DBLP:conf/sigir/VulicM15}, and cross-lingual plagiarism detection \cite{potthast2011cross,FrancoSalvador2016:KNOSYS}. The recently proposed CL STS methods \cite{brychcin-svoboda:2016:SemEval,jimenez:2016:SemEval} are, however, not intrinsically cross-lingual because they first employ a full-blown MT systems to translate one sentence of each pair and then apply existing monolingual STS models. Since creation of parallel corpora for MT is arguably the most obvious application of a cross-lingual STS system, using an MT system to build a cross-lingual STS model seems like introducing a circular (i.e., ``chicken or the egg'') problem. On top of that, open and robust MT models still do not exist for many language pairs, which further impedes the wide applicability of the proposed CL STS models.
Most of the monolingual STS models (most of which are for English) are supervised \citep[\textit{inter alia}]{vsaric2012takelab,bar2012ukp,hanig2015exb} and focus primarily on learning the best combination of numerous features for a non-linear regression model. But even the rare, yet quite successful, unsupervised models \cite{kashyap2014meerkat,sultan2014dls} rely on various language-specific tools (e.g., named entity recognition, dependency parsing) and resources, e.g., WordNet \cite{fellbaum98}. The fact that such tools and resources exist only for a handful of languages also limits the impact of these models.

In contrast, in this work we present a resource-light approach to cross-lingual STS that can easily be applied to a wide range of language pairs. The main contributions of this work are as follows: 

\begin{enumerate}

\item An unsupervised model for cross-lingual STS which requires no language-specific tools and resources. The model merely relies on (1) the availability of large corpora for both input languages, which is satisfied even for severely under-resourced languages \cite{de2002web,ljubevsic2011hrwac}, and (2) a set of word translation pairs of limited size;\alertzwei{\footnote{\alertzwei{Most recently, \citet{artetxe-labaka-agirre:2017:Long} have shown that a shared bilingual embedding space of comparable quality can be induced even without any manually created word translation pairs.}}}

\item An extensive intrinsic evaluation of the proposed model on three benchmark STS datasets and for three different language pairs, one of which includes an under-resourced language;

\item Extrinsic evaluations of the proposed model on two different tasks: aligning parallel sentences from comparable corpora for machine translation and cross-lingual plagiarism detection. To the best of our knowledge, these are the first extrinsic evaluations of a cross-lingual STS model. 
\end{enumerate}

\noindent In our approach, we first independently train word embeddings for each language and then project vectors from one language (i.e., the source language) to the embedding space of the second language (i.e., the target language) using the linear \textit{translation matrix} model proposed by Mikolov et al.~(\citeyear{mikolov13}). We then measure the semantic similarity between short texts of different languages with three different unsupervised similarity scores, each of which exploits the word vectors in the shared bilingual embedding space. We first evaluate the proposed method on three different STS datasets and for three language pairs -- English-Spanish, English-Italian, and English-Croatian. The obtained experimental results show that the proposed approach (1) reaches performance levels reasonably close to those of supervised and resource-intensive STS models and (2) exhibits stable performance across all three language pairs (i.e., no significant performance drop when an under-resourced language like Croatian is involved). Finally, we evaluate our CL STS method on two extrinsic tasks: (1) parallel sentence alignment from comparable corpora \cite{smith2010extracting} (i.e., creation of training data for machine translation systems) and (2) cross-lingual plagiarism detection \cite{potthast2010evaluation}. Results from both extrinsic evaluations show that our unsupervised resource-light approach performs on par with significantly more complex and resource-intensive state-of-the-art models on both tasks.   
\section{Related Work}

The explosion of research on (monolingual) semantic textual similarity of short texts can be credited to the SemEval-2012 Pilot on Semantic Textual Similarity \cite{agirre2012semeval}, although preceding efforts exist \cite{islam2008semantic,oliva2011symss}. The best-performing systems in the first STS challenge \cite{vsaric2012takelab,bar2012ukp} were methodologically almost identical -- they combined many content comparison features -- ranging from simple ngram overlaps, over named entity alignments to similarity of latent representations -- with a supervised regression model. 

Later SemEval campaigns saw successful unsupervised STS models. Han et al.~(\citeyear{han2013umbc}) count pairs of semantically aligned words between the sentences, with the measure of similarity for two given words being based on their latent semantic vectors and relations between these words in WordNet. The similarity score of Sultan et al.~(\citeyear{sultan2014dls}) also counts the number of aligned word pairs, but they further employ named entity recognition and dependency parsing to align the words. Albeit unsupervised, these models use language-specific resources and tools (parsing, named entity recognition, WordNet), which prevents them from being directly applicable to under-resourced languages. 

\alert{The core STS task was extended to include the interpretability of similarity scores \cite{agirrea2015semeval,lopez2017interpretable}. The goal of the interpretable STS  is to provide reasoning behind the assigned similarity scores by identifying the alignment between pairs of segments across the two sentences, assigning to each alignment a relation type and a similarity score. At the moment, we focus only on the core STS task in the cross-lingual setting, and leave the intepretability for future work.}    

Although there are many applications that would directly benefit from systems for detecting semantic similarity between short texts across languages (e.g., machine translation, plagiarism detection, cross-lingual retrieval), the cross-lingual setting has been considered only in the most recent edition of the SemEval STS task \cite{agirre-EtAl:2016:SemEval1} and only for a single language pair (English-Spanish). Although the best performing systems \cite{brychcin-svoboda:2016:SemEval,jimenez:2016:SemEval} produce the similarity scores that have 90\% correlation with human scores, these systems are not truly cross-lingual because they merely use full-blown MT systems (e.g., Google translate) to translate one of the texts into another language which then allows them to apply resource-intensive monolingual alignment models, such as the one of Sultan et al.~(\citeyear{sultan2014dls}). These systems thus have limited applicability as they cannot be used for pairs of languages for which no robust MT model exists. 

Cross-lingual methods have also been proposed for a closely related task of recognizing textual entailment. However, the CL RTE approaches are also dominantly supervised, also employ full-blown MT systems \cite{turchi2013altn}, and rely on lexico-semantic resources \cite{castillo2011wordnet}, all of which prevent their off-the-shelf application to an arbitrary pair of languages.       

In contrast to all above-mentioned methods, we present the CL STS approach that is unsupervised and does not require any language-specific resources and tools. For a given pair of languages, our method requires only (1) a reasonably large corpora for each of the languages and (2) a reasonably small set of word translation pairs. This allows for a wide applicability of the proposed cross-lingual STS approach.    
 
\section{Resource-Light Cross-Lingual STS}

\begin{figure}
\begin{center}
\includegraphics[scale=0.55]{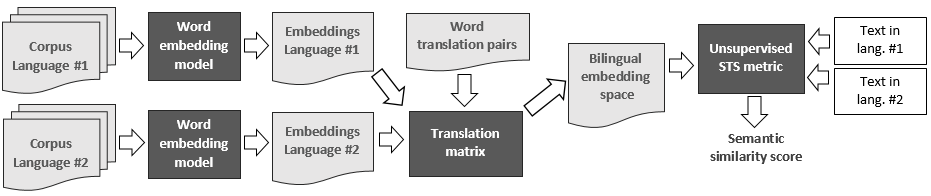}
\caption{Workflow of the resource-light approach for cross-lingual STS.}
\label{fig:workflow}
\end{center}
\end{figure}

Our cross-lingual STS model exploits the recently ubiquitous word embeddings \cite{mikolov2013distributed,pennington2014glove} and leverages the linear model for translating embedding spaces \cite{mikolov13}. The overall workflow of the proposed resource-light cross-lingual STS approach is given in Figure \ref{fig:workflow}. 
We start by independently training word embeddings for each of the two languages from their respective monolingual corpora. We then use a limited set of word translation pairs to compute the linear translation matrix that maps vectors from the embedding space of one language to the embedding space of the other language, effectively inducing this way the bilingual embedding space. Finally, having words of both languages represented with vectors of the joint bilingual embedding space allows us to compute resource-agnostic and unsupervised similarity scores by aligning words of different languages based on the similarities of their vectors.

\subsection{Building a Shared Embedding Space}

We first train monolingual word embeddings. Word embeddings are dense numeric vectors of words that have been shown to capture well the meaning (i.e., semantics) of words. These vectors can be learned from large corpora using different models, e.g., Skip-Gram \cite{mikolov2013distributed} or GloVe \cite{pennington2014glove}, that, one way or another, exploit co-occurrence information and contexts in which words appear. Independent training of monolingual embeddings, however, produces non-associated embedding spaces that do not contain similar vectors for the same concept across languages (e.g., the monolingual embedding vector for the English word \textit{``dog''} will not be similar to the embedding vector for its Italian translation \textit{``cane''}). 

In order to have the words from both languages in the same embedding space, we need to project the embedding space of one language into the embedding space of the other. Mikolov et al.~(\citeyear{mikolov13}) have shown that the mapping between the independently trained word embeddings is linear. This means that, given a limited number of word translation pairs, we can learn the translation matrix that translates embeddings from one embedding space to the other. Given the training set of word translation pairs (i.e., the set of paired monolingual embeddings), $\{s_{i},t_{i}\}^{n}_{i=1}$, $s_{i} \in \mathbb{R}^{d_{s}}$, $t_{i} \in \mathbb{R}^{d_{t}}$ (where $d_s$ is the dimension of the source language embeddings and $d_t$ is the dimension of the target language embeddings), we obtain the translation matrix $M \in \mathbb{R}^{d_{t} \times d_{s}}$ by minimizing the following sum:
\begin{equation*}
\sum^{n}_{i=1} \lVert Ms_{i} - t_{i} \rVert_{2}
\end{equation*}
The translation matrix $M$ is trained using a set of gold translation pairs which is very small in comparison to the size of the monolingual vocabularies. \alertzwei{Recently, \citet{artetxe-labaka-agirre:2017:Long} have shown that an equally robust bilingual embedding space can be induced even without any manually created word translation pairs.} Once obtained, the matrix $M$ is used to project the whole vocabulary of the source language to the embedding space of the target language.

\subsection{Computing Semantic Textual Similarity}

Joint bilingual embedding space allows for semantic comparison of words from different languages. For a given pair of words ($w_s$, $w_t$), $w_s$ from the source language and $w_t$ from the target language, we measure the semantic similarity simply as the cosine of the angle between their corresponding vectors in the bilingual embedding space: 
\begin{equation*}
\mathit{sim}(w_s,w_t) = \cos(Mv_s(w_s),v_t(w_t))
\end{equation*}
\noindent where $v_s(w_s)$ gets the initial monolingual embedding for the source word $w_s$ (which is then mapped to the target embedding space via the multiplication with the translation matrix $M$), and $v_t(w_t)$ is the monolingual embedding of the target word $w_t$. We next exploit this cross-lingual word similarity in three \alert{simple unsupervised} CL STS scores.\footnote{All three scores can be used for monolingual STS as well, in which case we do not need any mapping of the embedding spaces.} \alert{Aiming for resource-light and language-independent STS methodology, these similarity metrics have been designed to rely on similarities and word-alignments stemming exclusively from the semantic vectors of words, which can be obtained inexpensively for virtually any language. In the monolingual setting, for languages for which robust linguistic tools (like parsers and named entity recognizers) exist, alignment methods that additionally exploit these signals (parses and named entities), e.g., \citep{sultan2014dls}, will outperform our simple similarity measures. However, such methods cannot be applied in the cross-lingual setting unless one (or both) sentences are machine translated to a language for which necessary linguistic tools exist \citep{brychcin-svoboda:2016:SemEval}}.

\subsubsection{Greedy Association Similarity.}

The computation of the \textit{greedy association similarity} score is based on finding for each word from the source language text the most semantically similar word from the target language text, and vice versa. The greedy association allows the same word from the target language text (source language text) to be coupled with multiple source language words (target language words), assuming it is the most semantically similar target word (source word) for all of those source words (target words). Let $S$ be the bag of words of the short text in the source language, and $T$ be the bag of words of the target language text. Also, let $t(w_s)$ return the most semantically similar word from the target text for the given source word $w_s$ and let $s(w_t)$ find the most similar word from the source text for the given target word $w_t$. The non-symmetric greedy association scores are then computed as follows: 
\begin{align*}
\mathit{ngas}(S, T) & = \frac{1}{|S|}\sum_{i = 1}^{|S|}{\mathit{sim}(w_{s}^{i},t(w_{s}^{i}))},  \\
\mathit{ngas}(T, S) & = \frac{1}{|T|}\sum_{i = 1}^{|T|}{\mathit{sim}(w_{t}^{i},s(w_{t}^{i}))}.  \\
\end{align*}
\noindent      
The final (symmetric) greedy association similarity score is computed as the average of the two non-symmetric scores, i.e., $gas(S,T) = \frac{1}{2}\cdot(ngas(S,T) + ngas(T,S))$. To compute the greedy association score, we first must compute the similarity scores between all words from $S$ and all words from $T$ and then for each word from one set find the most similar word in the other set. Hence, the complexity of computing the greedy association score is $\mathcal{O}(|S|\cdot |T|)$.

\subsubsection{Optimal Alignment Similarity.}

Unlike for the greedy association similarity, for the computation of the \textit{optimal alignment similarity} score we are looking for an optimal alignment between the source and target words. The optimal alignment is the one for which the sum of similarities of aligned pairs is maximal. That is, we are looking for the optimal alignment $\{(w^{i}_{s}, w^{i}_{t})\}^{N}_{i = 1}$, where $w^{i}_{s}$ is a word from the source text bag of words $S$, $w^{i}_{t}$ is a word from the target text bag of words $T$, and $N$ is the number of aligned pairs, equal to the number of words in the shorter of the two texts. The optimal alignment is the one maximizing the sum of pairwise word similarities:
\begin{equation*}
\mathit{align}(S, T) = \max_{\{w^{i}_{s}, w^{i}_{t}\}_{i=1}^{N}}{\sum_{i=1}^{N}{\mathit{sim}(w^{i}_{s}, w^{i}_{t})}}
\end{equation*}  
\noindent We find the optimal alignment using the Kuhn-Munkres algorithm \cite{kuhn1955hungarian},\footnote{The algorithm is also known as the \textit{Hungarian algorithm}.} which provides the optimal solution to an alignment problem in polynomial time.\footnote{Although the originally proposed algorithm has the complexity of $\mathcal{O}(n^4)$, the improved variant of the algorithm has the cubic complexity $\mathcal{O}(n^3)$.} As the optimal alignment algorithm requires the same number of elements in both multisets, we pad the shorter of the texts with artificial tokens that are maximally dissimilar (i.e., have the similarity score of -1) to all words from the longer text. Because longer texts will have more aligned pairs and thus larger similarity score (on the account of length rather than actual similarity), we normalize the $\mathit{align}$ score with the length of each of the two sentences, and produce the average of the two normalized scores as the final optimal alignment similarity score:
\begin{equation*}
\mathit{oas}(S,T) = \frac{\mathit{align}(S, T)\cdot (|S| + |T|)}{2 \cdot|S|\cdot|T|}.
\end{equation*}
\noindent As the cubic time complexity of the Hungarian algorithm asymptotically dominates over the quadratic complexity of computing pairwise word similarities between input texts $S$ and $T$, the complexity of computing the optimal alignment similarity amounts to $\mathcal{O}(\max(|S|, |T|)^3)$.   

\subsubsection{Aggregation Similarity.} Unlike the previous two semantic similarity scores for which we explicitly compute the similarities for all pairs of words between the short texts, in order to find the greedy or optimal alignment, for the \textit{aggregation similarity} score we first compute the aggregate embeddings for each of the texts and then compare the similarity between these aggregate short text embeddings. The embeddings of the short source text $S$, denoted $v(S)$, and the embedding of the short target text $T$, denoted $v(T)$, are computed as follows: 
\begin{align*}
v(S) & = \frac{1}{|S|}\sum_{w_s \in S}{M v_s(w_s)}, \\ 
v(T) & = \frac{1}{|T|}\sum_{w_t \in T}{v_t(w_t)}.
\end{align*}
The aggregation similarity score is then computed as a cosine of the angle between the aggregate embedding vectors of the two short texts: 
\begin{equation*}
\mathit{agg}(S,T) = \cos(v(S), v(T)).      
\end{equation*}
Given that for computing the aggregation similarity between sentences we only have to sum the embeddings of words in each of the input sentences $S$ and $T$ and perform a single cosine computation (constant time, $\mathcal{O}(1)$) between aggregate embeddings, the computation of the aggregation similarity score has linear time complexity with respect to the length of the input texts, i.e., $\mathcal{O}(|S| + |T|)$.

\section{Intrinsic Evaluation}
\label{sec:eval}

We first evaluate the proposed resource-light cross-lingual STS metrics intrinsically on three different STS evaluation datasets and for three different language pairs: English-Spanish (EN-ES), English-Italian (EN-IT), and English-Croatian (EN-HR). We have selected Spanish because of a readily available cross-lingual English-Spanish STS dataset\footnote{\url{http://alt.qcri.org/semeval2016/task1/data/uploads/sts2016-cross-lingual-test.tar.gz}} from the SemEval 2016 STS task \cite{agirre-EtAl:2016:SemEval1}. Besides Spanish, we chose Italian because we had easy access to native speakers of this language. Finally, because we claim that the main advantage of our method is its applicability to resource-poor languages, we decided to have one such language in our evaluation. Same as for Italian, we chose Croatian because we had access to a native speaker of that language.   


\subsection{Embeddings and Translation Matrices}

To train the translation matrices, we must first obtain monolingual word embeddings for all four languages. In all experiments, we mapped the embeddings of the three other languages to the English embedding space. \alertzwei{For each of the four languages -- English, Spanish, Italian, and Croatian -- we trained both (1) the 300-dimensional word embeddings using the Continuous Bag of Words (CBOW) model \citep{mikolov2013distributed} and (2) the 300-dimensional word embeddings using the GloVe model \citep{pennington2014glove}.
Having produced word embeddings for all languages with two different methods allowed for different types of source-to-target embedding space mapping -- we tested (1) CBOW to CBOW, (2) CBOW to GloVe, and (3) GloVe to GloVe embedding space translations. This allowed insight into whether using different models for producing source and target monolingual embeddings affects the quality of the obtained translation matrices (i.e., the quality of the shared embedding space).} We trained the Spanish embeddings on the Spanish Billion Words (SBW) Corpus,\footnote{\url{http://crscardellino.me/SBWCE/}} Croatian embeddings on the hrWaC corpus \citep{ljubevsic2011hrwac}, whereas we used the Wikipedia dumps from the Polyglot project\footnote{\url{https://sites.google.com/site/rmyeid/projects/polyglot}} for English and Italian. The details of the monolingual word embeddings and the corpora on which they were trained are listed in Table \ref{tbl:embeddings}. 

\begin{table}
\centering
\begin{tabular}{l c c c c}
\toprule 
Language & Corpus & Size (in~tok.) & Method & Dims \\ \midrule
English & EN Wikipedia & 1.7B & GloVe & 300 \\
English & EN Wikipedia & 1.7B & CBOW & 300 \\ \midrule
Spanish & SBW & 1.5B & CBOW & 300 \\
Italian & IT Wikipedia & 0.3B & CBOW & 300 \\
Croatian & hrWaC & 1.2B & CBOW & 300 \\
\bottomrule
\end{tabular}
\caption{Details on the trained monolingual word embeddings.}
\label{tbl:embeddings}
\end{table}

\alert{Following previous studies on post-hoc mapping of monolingual embedding spaces \cite{mikolov13,dinu2014improving}, for the training of translation matrices, we used the word translation pairs consisting of most frequent words in one of the languages}. We selected the 4200 most frequent English words and translated them to all three other languages via Google translate.\footnote{\url{https://translate.google.com/}} We then had native speakers of target languages fix incorrect automatic translations. In each case, we discarded pairs where (1) an English word had a multi-word translation in the other language or (2) one of the words, English or the word of other language, was not in its respective vocabulary of word embeddings. This left us with sets of between 3500 and 3700 word translation pairs (depending on the language pair), 200 of which we always left for testing the translation quality. \alertzwei{However, the most recent research results \citep{artetxe-labaka-agirre:2017:Long} show that a reliable shared embedding space can be induced using only a handful of aligned embedding vectors (to which end the embeddings of, e.g., Arabic digits, present in almost every language, can be used). Thus, by using a method such as \citep{artetxe-labaka-agirre:2017:Long} to induce a shared embedding space, one can eliminate even this small manual effort of creating a few thousand word translation pairs.} 

We learned the optimal values of the translation matrices stochastically, using the Adam algorithm \cite{kingma2014adam}. The quality of the obtained translation matrices is shown in Table \ref{tbl:trmat} in terms of precision at rank one (P@1) and precision at rank five (P@5). These numbers reflect, respectively, the percentage of cases in which the correct translation of the English word from the test set was retrieved by the trained translation matrix as the most similar or among the five most similar words in the other language. \alert{For example, when we (1) translate the embedding vector of some Spanish word $w_{ES}$ using the learned Spanish-English (CBOW $\rightarrow$ GloVe) translation matrix and (2) rank all words from the English vocabulary (several hundred thousand entries) according to the similarity of their embedding vectors with the translated embedding of $w_{ES}$, we will find the English word $w_{EN}$ that is the dictionary translation of $w_{ES}$ within the 5 top-ranked English words in 66\% of the cases and as the top-ranked in 48\% of the cases.} Overall, the obtained levels of translation performance, shown in Table \ref{tbl:trmat}, are comparable to those reported in the original work \citep{mikolov13}.

\setlength{\tabcolsep}{4.5pt}
\begin{table}[t]
\centering
\begin{tabular}{l cc cc cc}
\toprule 
  & \multicolumn{2}{c}{CBOW$\rightarrow$CBOW} & \multicolumn{2}{c}{CBOW$\rightarrow$GloVe} & \multicolumn{2}{c}{\alertzwei{GloVe$\rightarrow$GloVe}} \\
\cmidrule(lr){2-3} \cmidrule(lr){4-5} \cmidrule(lr){6-7} 
Mapping & P@1~(\%) & P@5~(\%) & P@1~(\%) & P@5~(\%) & \alertzwei{P@1~(\%)} & \alertzwei{P@5~(\%)} \\ \midrule
ES$\rightarrow$EN  & 35.6 & 55.0 & 48.9 & 66.7 & \alertzwei{45.6} & \alertzwei{67.8} \\
IT$\rightarrow$EN & 28.4 & 53.1 & 35.5 & 57.9 & \alertzwei{32.0} & \alertzwei{55.5} \\
HR$\rightarrow$EN & 29.9 & 52.9 & 32.2 & 57.5 & \alertzwei{32.3} & \alertzwei{48.7} \\
\bottomrule
\end{tabular}
\caption{Evaluation of translation matrices.}
\label{tbl:trmat}
\end{table}
\setlength{\tabcolsep}{6pt}

\alert{Somewhat surprisingly, we consistently observe better translation quality when we map the CBOW embeddings of other languages to English GloVe embeddings than when we use CBOW embeddings for English as target as well. \alertzwei{Also, using CBOW embedding at source and GloVe embedding on target sizes produces slightly better translation performance than those obtained using GloVe embeddings at both sides of mapping}. Although one would perhaps intuitively expect the mapping to be better if both source and target embeddings were built using the same embedding method, this is not necessarily the case. Both GloVe and CBOW build embedding spaces with similar properties and linear relations between embedding vectors. However, the monolingual English GloVe embeddings seem to be of higher quality than the monolingual English CBOW embeddings. We verify this by measuring the performance of both GloVe- and CBOW-induced embedding spaces on the semantic similarity benchmark dataset SimLex-999 \citep{hill2015simlex} where GloVe-induced embedding space reaches 38.9\% Spearman correlation with gold similarity scores, compared to 33.5\% reached by CBOW embeddings. Following these results, in all STS and extrinsic evaluations that follow we built the translation matrices using the CBOW source embeddings (for other three languages) and GloVe English embeddings as the target embedding space.} 

\alert{We have also trained translation matrices using the word translation pairs randomly selected from the synsets from BabelNet, the large multilingual knowledge base \cite{navigli2012babelnet}. However, by using randomly sampled translation word pairs from BabelNet, we were consistently obtaining translation matrices of lower quality (5\% average performance drop in both P@1 and P@5) than when using word translation pairs containing the most frequent English words.}

The translation quality is slightly better for Spanish than for the other two languages. We speculate that this is \alert{due to combination of (1) Spanish SBW used to train Spanish monolingual embeddings being much larger than Italian Wikipedia and somewhat larger than hrWaC, used to train Italian and Croatian monolingual embeddings, respectively, and} (2) Spanish words being more frequent in English corpora than the words of the other two languages.

\subsection{Datasets} 

\alert{In order to enable insights into the suitability of proposed STS models on different types of short texts, we have decided to evaluate them on a battery of datasets, mutually different both in terms of genre and text length}: 
\begin{itemize}

\item \textsc{News-16} dataset is the readily available cross-lingual English-Spanish dataset used in the SemEval 2016 STS task, consisting of 301 pairs of news headlines \citep{agirre-EtAl:2016:SemEval1}. We translate the Spanish sentences into Italian and Croatian in order to create English-Italian and English-Croatian versions of the same dataset as well as into English in order to create the monolingual English version of the dataset;

\item \alert{\textsc{MulSrc-16} dataset is the readily available cross-lingual English-Spanish dataset used in the SemEval 2016 STS task, consisting of 294 sentences originating from multiple sources (new headlines, question-question and answer-answer pairs from a QA dataset, pairs of sentences from plagiarism detection datasets) \citep{agirre-EtAl:2016:SemEval1}. The sentences in the dataset are mutually much more heterogeneous than in the \textsc{News-16} dataset. Same as for \textsc{News-16} dataset, we translate the Spanish sentences into Italian, Croatian, and English in order to create English-Italian, English-Croatian, and monolingual English-English dataset variants;}

\item \alert{Microsoft Research video captions dataset (\textsc{MSRVid-12}) is a monolingual English dataset from the SemEval 2012 STS task \citep{agirre2012semeval}, consisting of 750 pairs of very short and grammatically simple video caption sentences. To create multilingual versions of this dataset, we translated one sentence of each pair to Spanish, Italian, and Croatian;}

\item  \alert{OntoNotes-WordNet dataset (\textsc{OnWN-12}), is a also monolingual English dataset from the SemEval 2012 STS task \citep{agirre2012semeval}, consisting of 750 pairs of concept definitions from OntoNotes and WordNet. To create multilingual versions of this dataset, we translated one sentence of each pair to Spanish, Italian, and Croatian. Definitional sentences of the \textsc{OnWN-12} dataset vary in length much more than the \textsc{MSRVid-12} dataset.}

\end{itemize}

For all of the dataset, we created the missing cross-lingual and monolingual variants by first performing automated machine translation with Google translate and then letting native speakers of target languages fix the errors introduced by machine translation.\footnote{\alertzwei{We make the cross-lingual datasets, along with word translation pairs and linear mapping code, freely available at \url{https://bitbucket.org/gg42554/cl-sts}}}    

\alert{We summarize the key information about the datasets in Table \ref{tbl:datasets}. We observe that the datasets are mutually very different in terms of average sentence length, with \textsc{News-16} dataset sentences being on average five times longer than the \textsc{MSRVid-12} sentences. \textsc{MulSrc-16} and \textsc{OnWN-12} datasets have larger within-dataset relative variance in sentence length than \textsc{News-16} and \textsc{MSRVid-12}. On top of high sentence-length variance, instances in \textsc{MulSrc-16} also vary in genre.     
We believe that evaluating STS models on the collection of datasets with such mutually differing properties (sentence length, genre, and both length- and genre-based homogeneity/heterogeneity can provide more insights into the strengths and weaknesses of different STS algorithm variants.}       
\setlength{\tabcolsep}{5pt}
\begin{table}
\begin{center}
\small{
\begin{tabular}{l c cccc}
\toprule
 & & \multicolumn{4}{c}{Average sentence length} \\
\cmidrule(lr){3-6}
Dataset & Num.~pairs & EN & ES & IT & HR \\ \midrule
\textsc{News-16} & 301 & 29.2 $\pm$ 10.4 & 32.7 $\pm$ 11.3 & 31.0 $\pm$ 10.9 & 25.8 $\pm$ 9.2 \\ 
\textsc{MulSrc-16} & 294 & 12.6 $\pm$ 7.3 & 14.0 $\pm$ 8.3 & 13.3 $\pm$ 7.7 & 11.1 $\pm$ 6.6 \\
\textsc{MSRVid-12} & 750 & 6.6 $\pm$ 1.7  & 7.2 $\pm$ 2.2 & 6.9 $\pm$ 2.0 & 4.6 $\pm$ 1.6 \\ 
\textsc{OnWN-12} & 750 & 8.1 $\pm$ 4.4 & 7.7 $\pm$ 3.8 & 7.2 $\pm$ 3.5 & 5.8.0 $\pm$ 2.8 \\ 
\bottomrule
\end{tabular}
}
\caption{Datasets used for intrinsic STS evaluation}
\label{tbl:datasets}
\end{center}
\end{table}

\subsection{Results and Discussion}

We evaluate all three of our cross-lingual STS scores -- greedy association similarity (\textsc{GrAssoc}), optimal alignment similarity (\textsc{OptAlign}) and aggregation similarity (\textsc{Aggreg}) on the cross-lingual datasets created as described above. Besides the cross-lingual evaluation, for the sake of comparison, we evaluated the same similarity scores on the original monolingual variants of the (\textsc{MSRVid-12}) and (\textsc{OnWN-12}) datasets. The model performance, in terms of Pearson correlation ($\rho$) with human-assigned similarity scores is shown in Tables \ref{tbl:resultssts16} and \ref{tbl:resultssts12}. In Table \ref{tbl:resultssts16}, we compare the performance on the English-Spanish variant of the \textsc{News-16} dataset with the best performing models from the SemEval 2016 STS task \cite{brychcin-svoboda:2016:SemEval,jimenez:2016:SemEval}. Similarly, in Table \ref{tbl:resultssts12}, we provide the performance of the best-performing monolingual STS models from SemEval 2012 \cite{vsaric2012takelab,bar2012ukp} \alert{as well as the performance of the state-of-the-art unsupervised model of \citet{sultan2014dls}} on the \textsc{OnWN-12} and \textsc{MSRvid-12} datasets. Although we are aware that the performances on monolingual and cross-lingual variants of datasets cannot be directly compared, we believe that this still provides useful context for interpreting the cross-lingual performance.
\setlength{\tabcolsep}{5.5pt}
\begin{table}[t]
\centering
\small{
\begin{tabular}{l cccc cccc}
\toprule 
& \multicolumn{4}{c}{\textsc{News-16}~(EN--)} & \multicolumn{4}{c}{\textsc{MulSrc-16}~(EN--)} \\ \cmidrule(lr){2-5} \cmidrule(lr){6-9} 
Model & EN & ES & IT & HR & EN & ES & IT & HR \\ \midrule
\textsc{OptAlign} & 90.8 & 86.6 & 84.8 & 78.4 & 82.6 & 77.2 & 70.4 & 64.8 \\
\textsc{GrAssoc} & 90.4 & 84.2 & 78.1 & 77.4 & 81.2 & 76.0 & 62.7 & 59.0 \\
\textsc{Aggreg} & 77.7 & 59.4 & 52.0 & 53.6 & 77.0 & 52.3 & 40.3 & 39.4  \\ \midrule 
Brychc\'{i}n,~Svoboda~(\citeyear{brychcin-svoboda:2016:SemEval}) & -- & \textbf{90.6} & -- & -- & -- & \textbf{81.9} & -- & -- \\
Jimenez~(\citeyear{jimenez:2016:SemEval}) & -- & 88.7 & -- & -- & -- & 81.8 & -- & --  \\
\bottomrule
\end{tabular}
}
\caption{Results on the \textsc{News-16} and \textsc{MulSrc-16} datasets ($\rho$, in \%).}
\label{tbl:resultssts16}
\end{table}
\setlength{\tabcolsep}{7.2pt}
\begin{table}[t]
\centering
\small{
\begin{tabular}{l cccc cccc}
\toprule 
& \multicolumn{4}{c}{\textsc{MSRvid-12}~(EN--)} & \multicolumn{4}{c}{\textsc{OnWN-12}~(EN--)} \\ \cmidrule(lr){2-5} \cmidrule(lr){6-9} 
Model & EN & ES & IT & HR & EN & ES & IT & HR \\ \midrule
\textsc{OptAlign} & 75.2 & 61.3 & 60.2 & 52.8 & 70.2 & 49.6 & 45.2 & 39.0 \\
\textsc{GrAssoc} & 75.4 & 59.7 & 53.8 & 48.8 & 68.2 & 49.0 & 44.3 & 38.5 \\
\textsc{Aggreg} & 76.7 & 59.9 & 50.1 &  49.5 & 66.6 & 39.0 & 31.2 & 29.3 \\ \midrule 
\citet{sultan2014dls} & 82.0 & --  & --  & -- & \textbf{72.3} & --  & --  & --  \\ \midrule
\citet{vsaric2012takelab} & \textbf{88.0} & -- & -- & -- & 70.5 & -- & -- & -- \\
\citet{bar2012ukp} & 87.4 & -- & -- & -- & 66.5 & -- & -- & -- \\
\bottomrule
\end{tabular}
}
\caption{Results on the \textsc{MSRvid-12} and \textsc{OnWN-12} datasets ($\rho$, in \%).}
\label{tbl:resultssts12}
\end{table}

The optimal alignment model (\textsc{OptAlign}), quite expectedly, consistently outperforms the greedy association model (\textsc{GrAssoc}). The \textsc{OptAlign} model in most cases also outperforms the corresponding aggregation similarity model (\textsc{Aggreg}) model. \alert{However, the gap in performance in favor of \textsc{OptAlign} is drastically wider on the two longer-sentence datasets, \textsc{News-16} and \textsc{MulSrc-16} (around 25\% on average) than on the two datasets with shorter sentences, \textsc{MSRvid-12} and \textsc{OnWN-12} (2-14\%  in cross-lingual settings). We believe that this is because the sentence embeddings averaged from a large number of word embeddings, as computed by the \textsc{Aggreg} model on long sentences, do not accurately capture the meaning of the sentence. Put differently, the aggregated sentence embeddings are semantically more accurate for shorter sentences, and fuzzier for longer sentences.} 

There is a significant performance drop between the monolingual models (EN-EN) and their respective cross-lingual models (EN-ES, EN-IT, and EN-HR), \alert{ranging from 5\% to 30\% (depending on the dataset and the cross-lingual language pair)}, that clearly shows how much imperfect embedding space translation affects the STS performance, as those performance drops cannot be credited to anything else. The performance for English-Spanish pair is consistently better than for the other two pairs, which more or less exhibit comparable performance. We believe that this merely reflects the differences in quality of the respective translation matrices (see Table \ref{tbl:trmat}). 

\alert{The performance of the unsupervised \textsc{OptAlign} model on the monolingual EN-EN variant of the \textsc{OnWN-12} dataset is comparable or better than that of best-performing supervised models \cite{vsaric2012takelab,bar2012ukp} and only 2\% worse than the state-of-the-art unsupervised model of \citet{sultan2014dls}}. We find this rather encouraging, considering that all of these models employ a range of tools and resources (e.g., parsing, named entity recognition, WordNet) for feature computation \alert{or finding optimal word alignments. On the other hand, on the original monolingual \textsc{MSRvid-12} dataset, \textsc{OptAlign} is outperformed by the resource-intensive model of \citet{sultan2014dls} by 7\%. This is because the similarities between short and grammatically simple sentences of \textsc{MSRvid-12} highly depend on the syntactic roles the concepts occupy (e.g., \textit{``The man is playing with the dog''} vs. \textit{``The dog is playing with the man''}). Whereas the model of \citet{sultan2014dls}, using dependency parsing between sentences can take syntactic roles of concepts into account to better align words between sentences, our simple STS metric -- designed to be language-independent and resource-light -- cannot.}        

The cross-lingual performances of the \textsc{OptAlign} model on the English-Spanish variants of the \textsc{News-16} and \textsc{MulSrc-16} datasets come reasonably close to the best performing models \cite{brychcin-svoboda:2016:SemEval,jimenez:2016:SemEval} from the SemEval 2016 STS cross-lingual evaluation. The 4-5\% difference in performance (86\% compared to 90\% on \textsc{News-16} and 77\% compared to 82\% on \textsc{MulSrc-16}) seems rather satisfying considering that \textsc{OptAlign} is very resource-light and language-independent, whereas the best-performing system \cite{brychcin-svoboda:2016:SemEval} employs a full-blown MT system (Google translate), and a word alignment model that requires a dependency parser, a named entity recognizer, and a paraphrase database \citep{sultan2014dls}. 

\subsection{Analysis of the Translation Set Size} 

\begin{figure*}[t!]
	\begin{center}
        \begin{subfigure}[b]{0.45\textwidth}
                \centering
				\includegraphics[scale=0.35]{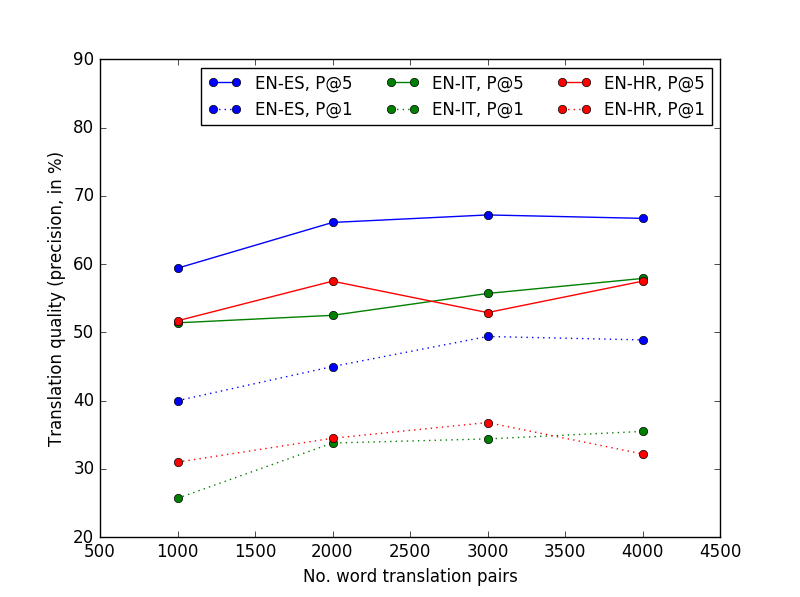}\\[0.5em]
                \caption{Translation matrix quality}
                \label{fig:transquality}
        \end{subfigure}\quad\   
        ~ 
        \begin{subfigure}[b]{0.45\textwidth}
                \centering
				\includegraphics[scale=0.35]{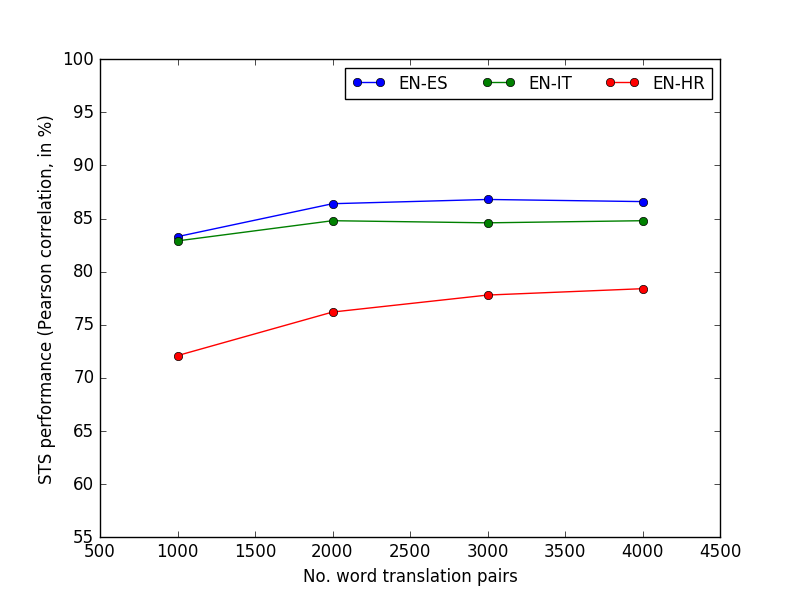} \\ [0.5em]
                \caption{Cross-lingual STS performance}
                \label{fig:sizests}
        \end{subfigure}
        \caption{Effects of the translation training set size.}
        \label{fig:graphs}
	\end{center}
\end{figure*}

Besides the sufficiently large corpora in both languages, the only resource our models need is the set of word translation pairs as the training set for the translation matrix. With the goal of the wide applicability in terms of different language pairs in mind, we aim to make our cross-lingual STS models as resource-light as possible. To that effect, we examine how the number of word translation pairs affects the quality of the obtained translation matrices and to which extent we may reduce the number of word translation pairs without significantly reducing the CL STS performance. To this end, we conducted experiments using sets of word translation pairs of four different sizes: 1K, 2K, 3K, and 4K. For each of the four translation sets and for each of the three language pairs, we evaluated (1) the quality of the obtained translation matrix (on the same test set of 200 translation pairs) and (2) the cross-lingual STS performance on the \textsc{News-16} dataset of our best-performing \textsc{OptAlign} model when using the corresponding translation matrix. These results are shown in Figure \ref{fig:graphs}.

These experiments show that we already reach stable cross-lingual STS performance by using 2K translation pairs for  the translation matrix optimization. The drop in performance compared to using 3K or 4K training pairs is small both in terms translation matrix quality STS performance. Moreover, for the language pair English-Spanish we observe no drop in performance when reducing the number of the word translation pairs from 4K to 2K. The more prominent drop in performance for all three language pairs is observed only when further reducing the number of word translation pairs from 2K to 1K. These results are encouraging as they indicate that we do not need large sets of word translation pairs to train translation matrices that are sufficiently good to allow stable CL STS performance.     

\section{Applications}

A cross-lingual STS system is useful for a range of tasks that require identifying short texts of similar meaning across languages. Two prominent such tasks on which we extrinsically evaluate our resource-light CL STS model are: (1) the parallel sentence extraction from comparable corpora \cite{smith2010extracting}, as parallel corpora are essential for training MT models; and (2) cross-lingual plagiarism detection \cite{potthast2011cross}. 

\subsection{Parallel Sentences Alignment from Comparable Documents}

As creating parallel corpora amounts to recognizing sentences in different languages with the same meaning (i.e., direct translations), we chose it as one of the two tasks on which to extrinsically evaluate our resource-light CL STS model. To this end, we used the English-Spanish portion of the dataset created by Smith et al.~(\citeyear{smith2010extracting}) consisting of twenty aligned Wikipedia articles (i.e., twenty pairs of comparable documents). 

We pair all English sentences with all Spanish sentences for each pair of comparable documents and compute the similarity scores using our best-performing \textsc{OptAlign} English-Spanish model. We then simply declare pairs of sentences with the similarity score above the threshold $\tau$ to be parallel sentences. Obviously, the value of the threshold $\tau$ regulates the precision-recall trade-off because lower $\tau$ values lead to higher recall and lower precision, whereas the higher $\tau$ values lead to higher precision and lower recall. In order to allow for a direct comparison with the supervised model of Smith et al.~(\citeyear{smith2010extracting}), we do not optimize the threshold $\tau$ in terms of, e.g., $F_1$ score, but rather evaluate the performance in terms of the average precision (AP), recall at the precision of 90\% (R@90), and recall at the precision of 80\% (R@80), as proposed by Smith et al.~(\citeyear{smith2010extracting}).

The results of the parallel sentence extraction evaluation are shown in Table \ref{tbl:extrinsic}. 
\begin{table}
\centering
\begin{tabular}{lccc}
\toprule 
Method & AP~(\%) & R@90~(\%) & R@80~(\%) \\ \midrule
\textsc{OptAlign} & 94.2 & 87.1  & 90.4 \\
Smith et al.~(\citeyear{smith2010extracting}) & 94.7 & 87.6 & 90.2 \\
\bottomrule
\end{tabular}
\caption{Performance on the task of parallel sentence extraction from comparable documents.}
\label{tbl:extrinsic}
\end{table}
The results show that our simple cross-lingual STS model reaches about the same performance level as the state-of-the-art model of Smith et al.~(\citeyear{smith2010extracting}). This result is quite encouraging given that our unsupervised and resource-light model matches the performance of the supervised model that (1) exploits out-of-domain seed parallel sentences for training the word alignment model, (2) employs a wide set of features, some of which are domain-specific (e.g., based on Wikipedia markup), and (3) exploits the fact that parallel sentences come in clusters and accordingly uses a supervised sequence labeling model to find the globally optimal sentence alignment \cite{smith2010extracting}.  

\subsection{Cross-Lingual Plagiarism Detection}

The second task on which the CL STS methods can be directly applied is the task of cross-lingual plagiarism detection. We compare the performance of our best-performing CL STS model with several state-of-the-art methods for cross-lingual plagiarism detection \cite{potthast:2011}. 

\subsubsection{Evaluation Setting}

Given a suspicious document $d_L$ in a language $L$ and a collection of source documents $D'_{L'}$ in a language $L'$, the task is to identify all fragments of $d_L$ that have been plagiarized from any of the documents from the collection $D'_{L'}$. 
%
In order to detect the plagiarism candidates between two documents $d_L$ and $d'_{L'}$ written in different languages $L$ and $L'$, the documents are first segmented into sets of fragments $\text{FC} \in d_L$ and $\text{FC}' \in d_L'$. Next, a similarity model $S$ is used to measure cross-lingual similarities $\text{SF}=\{\text{S}(F,F')\}$ between all pairs of text fragments $(F,F')$, $F \in \text{FC}$ and $F' \in \text{FC}'$. We evaluate our best-performing CL STS model (i.e., \textsc{OptAlign}) by plugging it into this framework as the similarity function $S$. 



To evaluate different similarity models for cross-lingual plagiarism detection, we use the Spanish-English (ES-EN) partition of the PAN-PC-11 dataset that was created for the 2011 plagiarism detection competition of PAN at CLEF.\footnote{\url{http://www.uni-weimar.de/en/media/chairs/webis/corpora/corpus-pan-pc-11/}}\textsuperscript{,}\footnote{\url{http://www.clef-initiative.eu/}} 
The plagiarism cases in PAN-PC-11 were generated using translation obfuscation with Google translate. In addition, the dataset also contains cases of plagiarism obtained with manual obfuscation following the automatic translation.

\subsubsection{Competing Models}

We compare our best performing cross-lingual STS model (according to intrinsic evaluation, this is the optimal alignment model), which we dub \textsc{CL-STS-OptAlign} in this evaluation, with three state-of-the-art similarity scores for plagiarism detection. The first similarity measure, named \textsc{CL-VSM}, is the cosine between TD-IDF weighted vector space model vectors of the text fragments. As the fragments are in different languages ($L$ and $L'$), each document vector $d_L$ is transformed into the bilingual form $d_{LL'}$ by concatenating the vector $d_{L'}$ which contains translations of words from $d_L$ obtained using a statistical dictionary. The statistical dictionary necessary for this model is trained using the word-alignment machine translation model IBM M1~\cite{och:2003} on the parallel JRC-Acquis~\cite{steinberger:2006} corpus. This similarity score is resource-intensive as it requires an existence of  a word alignment model for the two given languages, which in turns requires the existence of sufficiently large parallel corpora.   

The most similar method to ours is the Continuous Word Alignment-based Similarity Analysis (\textsc{CWASA}) model~\cite{FrancoSalvador2016:KNOSYS} which also measures the similarity by aligning words from two text fragments based on their continuous vectors. However, unlike CL-STS for which we obtain bilingual embedding vectors via the translation matrix model with only a limited number of word pairs, CWASA learns the bilingual embedding vectors by running the Siamese Neural Network (\textsc{S2Net}) model \cite{yih:2011} on the EN-ES DGT translation memory parallel corpora,\footnote{\url{https://ec.europa.eu/jrc/en/language-technologies/dgt-translation-memory}} which is drastically more difficult and expensive to obtain than a couple of thousands of word translation pairs. Due to the computational time of S2Net, we follow the setting from ~\cite{FrancoSalvador2016:KNOSYS} and restrict its vocabulary to the 20,000 most frequent words.

Finally, the Knowledge-Based document similarity (\textsc{KBSim}) model is based on grounding the content of text fragments in the large multilingual knowledge graph \cite{francosalvador-rosso-navigli:2014:EACL,FrancoSalvador2016:IPM,FrancoSalvador2016:KNOSYS}, namely BabelNet \cite{navigli2012babelnet}. The subgraph of BabelNet is extracted for each of the two fragments in different languages and the similarity between the corresponding extracted BabelNet subgraphs is used for plagiarism detection. The graph-based similarity score is further dynamically combined with the CL-VSM similarity, with weights assigned to each of the similarity components depending on the connectivity of the knowledge base subgraphs extracted for input text fragments. Besides the resources required by \textsc{CL-VSM}, \textsc{KBSim} additionally requires access to a multilingual knowledge base and to a part-of-speech tagger with lemmatization.  


\subsubsection{Evaluation Metrics and Results}


We measure the performance of different similarity scores with the overall document recall of plagiarized text measured at the character level. This experiment shows the potential of the models to detect cases of plagiarism at the document level, without differentiating between individual plagiarism cases. The character-level recall (R@\textit{k}) is measured using the top \textit{k} fragments from the source documents most similar to the fragment of the suspicious text. The results of the plagiarism detection evaluation are given in Table \ref{CLPD_expA1}. Our simple unsupervised and resource-light CL STS model performs on par with state-of-the-art similarity scores for cross-lingual plagiarism detection. \textsc{CL-STS-OptAlign} significantly outperforms the \textsc{CL-VSM} and the \textsc{CWASA} (\textsc{S2Net}) models for R@1 and \textsc{CL-VSM} for R@5. Although \textsc{KBSim} similarity numerically outperforms the \textsc{CL-STS-OptAlign} model for all four metrics, the differences are not statistically significant.\footnote{We tested all significances using the non-parametric bootstrap resampling \cite{efron1994introduction} test with $\alpha$ of 0.05 and 1000 samplings.} This findings are very promising as they show that robust detection of cross-lingual plagiarism can be achieved without expensive language-specific resources and tools. 
\begin{table}[!t]
  \begin{center}
    \scalebox{1.0}{
      \begin{tabular}{lllll}
      \toprule
	  Model&  R@1 & R@5 & R@10 & R@20~ \\ 
	  \toprule
	   \textsc{CL-VSM} & 0.791 & 0.880 & 0.905 & 0.924 \\
       \textsc{CWASA} (\textsc{S2Net}) & 0.859 & 0.909 & 0.921 & 0.936 \\ 
	   \textsc{KBSim} (\textsc{CL-VSM}) & 0.927 & 0.955 & 0.961 & 0.965 \\ \midrule
	   \textsc{CL-STS-OptAlign} & 0.895 & 0.930 & 0.940 & 0.948 \\
    \bottomrule
      \end{tabular}
    }  
  \end{center}
  \caption{ \label{CLPD_expA1} Performance analysis on the task of cross-lingual plagiarism detection (in terms of R@\textit{k}, where \textit{k} = \{1, 5, 10, 20\}).} 
\end{table} 
\section{Conclusion}

In this article, we presented an unsupervised resource-light approach to cross-lingual STS based on linear translations between monolingual embedding spaces. Unlike existing STS models (monolingual and cross-lingual alike), our method does not exploit expensive-to-build language-specific resources. Instead, our models require only large corpora for both input languages and a small set of word translation pairs -- resources that are cheap and easy to obtain for the vast majority of languages and language pairs. 

In the proposed approach we first construct the shared bilingual continuous vector space by mapping embedding vectors of words from one language to the embedding space of the other language. The mapping is achieved through the linear translation matrix which is learned using the set of aligned word pairs between the languages. We show that as few as 2000 word translation pairs is already enough to obtain high-quality translation matrices, leading to stable cross-lingual STS performance. 

We thoroughly evaluated the proposed CL STS model: (1) intrinsically on three STS datasets and for three different language pairs and (2) extrinsically on two different tasks: parallel sentence alignment from comparable corpora and CL plagiarism detection. The results of the intrinsic evaluation show that the resource-light CL STS exhibits competitive performance as well as stability across different language pairs, including the pair with Croatian as an under-resourced language. Both extrinsic evaluations reveal that our resource-light method performs on par with state-of-the-art models for respective tasks, which are, without exception, much more complex and resource-intensive.

Our future efforts will go in three different directions. First, we intend to investigate methods for constructing multilingual semantic spaces that either require no (or fewer) word translation pairs or produce higher quality mappings than the linear translation matrix model. Secondly, we intend to build CL STS models for other language pairs, to allow for wider adoption of CL STS in various tasks. Finally, we intend to evaluate the CL STS models in other extrinsic tasks such as cross-lingual document and passage retrieval or cross-lingual text classification.

%



\section*{References}
    \bibliographystyle{elsarticle-harv} 
    \bibliography{references.bib}

\begin{thebibliography}{48}
\expandafter\ifx\csname natexlab\endcsname\relax\def\natexlab#1{#1}\fi
\expandafter\ifx\csname url\endcsname\relax
  \def\url#1{\texttt{#1}}\fi
\expandafter\ifx\csname urlprefix\endcsname\relax\def\urlprefix{URL }\fi

\bibitem[{Agirre et~al.(2015)Agirre, Banea, Cardiec, Cerd, Diabe,
  Gonzalez-Agirrea, Guof, Lopez-Gazpioa, Maritxalara, Mihalcea,
  et~al.}]{agirrea2015semeval}
Agirre, E., Banea, C., Cardiec, C., Cerd, D., Diabe, M., Gonzalez-Agirrea, A.,
  Guof, W., Lopez-Gazpioa, I., Maritxalara, M., Mihalcea, R., et~al., 2015.
  Semeval-2015 {T}ask 2: Semantic textual similarity, {E}nglish, {S}panish and
  pilot on interpretability. In: SemEval. ACL, pp. 252--263.

\bibitem[{Agirre et~al.(2016)Agirre, Banea, Cer, Diab, Gonzalez-Agirre,
  Mihalcea, Rigau, and Wiebe}]{agirre-EtAl:2016:SemEval1}
Agirre, E., Banea, C., Cer, D., Diab, M., Gonzalez-Agirre, A., Mihalcea, R.,
  Rigau, G., Wiebe, J., 2016. Semeval-2016 {T}ask 1: {S}emantic textual
  similarity, monolingual and cross-lingual evaluation. In: {S}em{E}val. ACL,
  pp. 497--511.

\bibitem[{Agirre et~al.(2012)Agirre, Diab, Cer, and
  Gonzalez-Agirre}]{agirre2012semeval}
Agirre, E., Diab, M., Cer, D., Gonzalez-Agirre, A., 2012. Semeval-2012 {T}ask
  6: {A} pilot on semantic textual similarity. In: SemEval. pp. 385--393.

\bibitem[{Artetxe et~al.(2017)Artetxe, Labaka, and
  Agirre}]{artetxe-labaka-agirre:2017:Long}
Artetxe, M., Labaka, G., Agirre, E., July 2017. Learning bilingual word
  embeddings with (almost) no bilingual data. In: ACL. pp. 451--462.

\bibitem[{Aziz and Specia(2011)}]{aziz2011fully}
Aziz, W., Specia, L., 2011. Fully automatic compilation of
  {P}ortuguese-{E}nglish and {P}ortuguese-{S}panish parallel corpora. In: STIL.
  pp. 234--238.

\bibitem[{B{\"a}r et~al.(2012)B{\"a}r, Biemann, Gurevych, and
  Zesch}]{bar2012ukp}
B{\"a}r, D., Biemann, C., Gurevych, I., Zesch, T., 2012. {UKP}: Computing
  semantic textual similarity by combining multiple content similarity
  measures. In: SemEval. ACL, pp. 435--440.

\bibitem[{Bowman et~al.(2015)Bowman, Angeli, Potts, and
  Manning}]{bowman-EtAl:2015:EMNLP}
Bowman, S.~R., Angeli, G., Potts, C., Manning, C.~D., September 2015. A large
  annotated corpus for learning natural language inference. In: Proceedings of
  the 2015 Conference on Empirical Methods in Natural Language Processing.
  Association for Computational Linguistics, Lisbon, Portugal, pp. 632--642.

\bibitem[{Brychc\'{i}n and Svoboda(2016)}]{brychcin-svoboda:2016:SemEval}
Brychc\'{i}n, T., Svoboda, L., 2016. {UWB} at semeval-2016 {T}ask 1: {S}emantic
  textual similarity using lexical, syntactic, and semantic information. In:
  {S}em{E}val. ACL, pp. 588--594.

\bibitem[{Castillo(2011)}]{castillo2011wordnet}
Castillo, J.~J., 2011. A {W}ord{N}et-based semantic approach to textual
  entailment and cross-lingual textual entailment. International Journal of
  Machine Learning and Cybernetics 2~(3), 177--189.

\bibitem[{Dagan et~al.(2010)Dagan, Dolan, Magnini, and
  Roth}]{dagan2010recognizing}
Dagan, I., Dolan, B., Magnini, B., Roth, D., 2010. Recognizing textual
  entailment: {R}ationale, evaluation and approaches. Natural Language
  Engineering 16~(1), 105--105.

\bibitem[{De~Schryver(2002)}]{de2002web}
De~Schryver, G.-M., 2002. Web for/as corpus: A perspective for the african
  languages. Nordic Journal of African Studies 11~(2), 266--282.

\bibitem[{Dinu et~al.(2015)Dinu, Lazaridou, and Baroni}]{dinu2014improving}
Dinu, G., Lazaridou, A., Baroni, M., 2015. Improving zero-shot learning by
  mitigating the hubness problem. In: ICLR Workshop papers.

\bibitem[{Efron and Tibshirani(1994)}]{efron1994introduction}
Efron, B., Tibshirani, R.~J., 1994. An introduction to the bootstrap. CRC
  press.

\bibitem[{Fellbaum(1998)}]{fellbaum98}
Fellbaum, C. (Ed.), 1998. Word{N}et: An Electronic Lexical Database. MIT Press,
  Cambridge, Mass.

\bibitem[{Franco-Salvador et~al.(2016{\natexlab{a}})Franco-Salvador, Gupta,
  Rosso, and Banchs}]{FrancoSalvador2016:KNOSYS}
Franco-Salvador, M., Gupta, P., Rosso, P., Banchs, R.~E., 2016{\natexlab{a}}.
  Cross-language plagiarism detection over continuous-space- and knowledge
  graph-based representations of language. Knowledge-Based Systems 111, 87 --
  99.

\bibitem[{Franco-Salvador et~al.(2016{\natexlab{b}})Franco-Salvador, Rosso, and
  Montes~y G{\'o}mez}]{FrancoSalvador2016:IPM}
Franco-Salvador, M., Rosso, P., Montes~y G{\'o}mez, M., 2016{\natexlab{b}}. A
  systematic study of knowledge graph analysis for cross-language plagiarism
  detection. Information Processing \& Management 52~(4), 550--570.

\bibitem[{Franco-Salvador et~al.(2014)Franco-Salvador, Rosso, and
  Navigli}]{francosalvador-rosso-navigli:2014:EACL}
Franco-Salvador, M., Rosso, P., Navigli, R., 2014. A knowledge-based
  representation for cross-language document retrieval and categorization. In:
  EACL. ACL, pp. 414--423.

\bibitem[{Han et~al.(2013)Han, Kashyap, Finin, Mayfield, and
  Weese}]{han2013umbc}
Han, L., Kashyap, A., Finin, T., Mayfield, J., Weese, J., 2013. {UMBC}
  {EBIQUITY-CORE}: Semantic textual similarity systems. In: SemEval. pp.
  44--52.

\bibitem[{H{\"a}nig et~al.(2015)H{\"a}nig, Remus, and
  De~La~Puente}]{hanig2015exb}
H{\"a}nig, C., Remus, R., De~La~Puente, X., 2015. {ExB Themis}: Extensive
  feature extraction from word alignments for semantic textual similarity. In:
  SemEval. p. 264.

\bibitem[{Hill et~al.(2015)Hill, Reichart, and Korhonen}]{hill2015simlex}
Hill, F., Reichart, R., Korhonen, A., 2015. Simlex-999: Evaluating semantic
  models with (genuine) similarity estimation. Computational Linguistics
  41~(4), 665--695.

\bibitem[{Islam and Inkpen(2008)}]{islam2008semantic}
Islam, A., Inkpen, D., 2008. Semantic text similarity using corpus-based word
  similarity and string similarity. ACM Transactions on Knowledge Discovery
  from Data (TKDD) 2~(2), 10.

\bibitem[{Jimenez(2016)}]{jimenez:2016:SemEval}
Jimenez, S., 2016. {SERGIOJIMENEZ} at semeval-2016 {T}ask 1: Effectively
  combining paraphrase database, string matching, {W}ord{N}et, and word
  embedding for semantic textual similarity. In: {S}em{E}val. ACL, pp.
  749--757.

\bibitem[{Kashyap et~al.(2014)Kashyap, Han, Yus, Sleeman, Satyapanich, Gandhi,
  and Finin}]{kashyap2014meerkat}
Kashyap, A., Han, L., Yus, R., Sleeman, J., Satyapanich, T., Gandhi, S., Finin,
  T., 2014. {Meerkat Mafia}: Multilingual and cross-level semantic textual
  similarity systems. In: SemEval. pp. 416--423.

\bibitem[{Kingma and Ba(2014)}]{kingma2014adam}
Kingma, D., Ba, J., 2014. Adam: A method for stochastic optimization. arXiv
  preprint arXiv:1412.6980.

\bibitem[{Kuhn(1955)}]{kuhn1955hungarian}
Kuhn, H.~W., 1955. The hungarian method for the assignment problem. Naval
  research logistics quarterly 2~(1-2), 83--97.

\bibitem[{Ljube{\v{s}}i{\'c} and Erjavec(2011)}]{ljubevsic2011hrwac}
Ljube{\v{s}}i{\'c}, N., Erjavec, T., 2011. {hrWaC} and {slWaC}: Compiling web
  corpora for croatian and slovene. In: TSD. Springer, pp. 395--402.

\bibitem[{Lopez-Gazpio et~al.(2017)Lopez-Gazpio, Maritxalar, Gonzalez-Agirre,
  Rigau, Uria, and Agirre}]{lopez2017interpretable}
Lopez-Gazpio, I., Maritxalar, M., Gonzalez-Agirre, A., Rigau, G., Uria, L.,
  Agirre, E., 2017. Interpretable semantic textual similarity: Finding and
  explaining differences between sentences. Knowledge-Based Systems 119,
  186--199.

\bibitem[{Madnani et~al.(2012)Madnani, Tetreault, and Chodorow}]{madnani2012re}
Madnani, N., Tetreault, J., Chodorow, M., 2012. Re-examining machine
  translation metrics for paraphrase identification. In: NAACL-HLT. pp.
  182--190.

\bibitem[{Mehdad et~al.(2011)Mehdad, Negri, and Federico}]{mehdad2011using}
Mehdad, Y., Negri, M., Federico, M., 2011. Using bilingual parallel corpora for
  cross-lingual textual entailment. In: ACL. pp. 1336--1345.

\bibitem[{Mikolov et~al.(2013{\natexlab{a}})Mikolov, Le, and
  Sutskever}]{mikolov13}
Mikolov, T., Le, Q.~V., Sutskever, I., 2013{\natexlab{a}}. Exploiting
  similarities among languages for machine translation. CoRR abs/1309.4168.

\bibitem[{Mikolov et~al.(2013{\natexlab{b}})Mikolov, Sutskever, Chen, Corrado,
  and Dean}]{mikolov2013distributed}
Mikolov, T., Sutskever, I., Chen, K., Corrado, G.~S., Dean, J.,
  2013{\natexlab{b}}. Distributed representations of words and phrases and
  their compositionality. In: NIPS. pp. 3111--3119.

\bibitem[{Navigli and Ponzetto(2012)}]{navigli2012babelnet}
Navigli, R., Ponzetto, S.~P., 2012. Babelnet: The automatic construction,
  evaluation and application of a wide-coverage multilingual semantic network.
  Artificial Intelligence 193, 217--250.

\bibitem[{Negri et~al.(2012)Negri, Marchetti, Mehdad, Bentivogli, and
  Giampiccolo}]{negri2012semeval}
Negri, M., Marchetti, A., Mehdad, Y., Bentivogli, L., Giampiccolo, D., 2012.
  Semeval-2012 {T}ask 8: {C}ross-lingual textual entailment for content
  synchronization. In: SemEval. pp. 399--407.

\bibitem[{Och and Ney(2003)}]{och:2003}
Och, F.~J., Ney, H., 2003. A systematic comparison of various statistical
  alignment models. Computational Linguistics 29(1), 19--51.

\bibitem[{Oliva et~al.(2011)Oliva, Serrano, del Castillo, and
  Iglesias}]{oliva2011symss}
Oliva, J., Serrano, J.~I., del Castillo, M.~D., Iglesias, {\'A}., 2011.
  {SyMSS}: A syntax-based measure for short-text semantic similarity. Data \&
  Knowledge Engineering 70~(4), 390--405.

\bibitem[{Pennington et~al.(2014)Pennington, Socher, and
  Manning}]{pennington2014glove}
Pennington, J., Socher, R., Manning, C.~D., 2014. Glove: Global vectors for
  word representation. In: EMNLP. pp. 1532--1543.

\bibitem[{Potthast et~al.(2011{\natexlab{a}})Potthast, Barr{\'o}n-Cede{\~n}o,
  Stein, and Rosso}]{potthast2011cross}
Potthast, M., Barr{\'o}n-Cede{\~n}o, A., Stein, B., Rosso, P.,
  2011{\natexlab{a}}. Cross-language plagiarism detection. Language Resources
  and Evaluation 45~(1), 45--62.

\bibitem[{Potthast et~al.(2011{\natexlab{b}})Potthast, Eiselt,
  Barr{\'o}n-Cede{\~n}o, Stein, and Rosso}]{potthast:2011}
Potthast, M., Eiselt, A., Barr{\'o}n-Cede{\~n}o, A., Stein, B., Rosso, P.,
  2011{\natexlab{b}}. Overview of the 3rd int. competition on plagiarism
  detection. In: CLEF.

\bibitem[{Potthast et~al.(2010)Potthast, Stein, Barr{\'o}n-Cede{\~n}o, and
  Rosso}]{potthast2010evaluation}
Potthast, M., Stein, B., Barr{\'o}n-Cede{\~n}o, A., Rosso, P., 2010. An
  evaluation framework for plagiarism detection. In: COLING. ACL, pp.
  997--1005.

\bibitem[{Resnik and Smith(2003)}]{resnik2003web}
Resnik, P., Smith, N.~A., 2003. The web as a parallel corpus. Computational
  Linguistics 29~(3), 349--380.

\bibitem[{{\v{S}}ari{\'c} et~al.(2012){\v{S}}ari{\'c}, Glava{\v{s}}, Karan,
  {\v{S}}najder, and Ba{\v{s}}i{\'c}}]{vsaric2012takelab}
{\v{S}}ari{\'c}, F., Glava{\v{s}}, G., Karan, M., {\v{S}}najder, J.,
  Ba{\v{s}}i{\'c}, B.~D., 2012. {TakeLab}: Systems for measuring semantic text
  similarity. In: SemEval. pp. 441--448.

\bibitem[{Smith et~al.(2010)Smith, Quirk, and Toutanova}]{smith2010extracting}
Smith, J.~R., Quirk, C., Toutanova, K., 2010. Extracting parallel sentences
  from comparable corpora using document level alignment. In: HLT-NAACL. ACL,
  pp. 403--411.

\bibitem[{Socher et~al.(2011)Socher, Huang, Pennin, Manning, and
  Ng}]{socher2011dynamic}
Socher, R., Huang, E.~H., Pennin, J., Manning, C.~D., Ng, A.~Y., 2011. Dynamic
  pooling and unfolding recursive autoencoders for paraphrase detection. In:
  NIPS. pp. 801--809.

\bibitem[{Steinberger et~al.(2006)Steinberger, Pouliquen, Widiger, Ignat,
  Erjavec, Tufis, and Varga}]{steinberger:2006}
Steinberger, R., Pouliquen, B., Widiger, A., Ignat, C., Erjavec, T., Tufis, D.,
  Varga, D., 2006. {JRC-Acquis}: A multilingual aligned parallel corpus with
  +20 languages. In: LREC.

\bibitem[{Sultan et~al.(2014)Sultan, Bethard, and Sumner}]{sultan2014dls}
Sultan, M.~A., Bethard, S., Sumner, T., 2014. {DLS@CU}: Sentence similarity
  from word alignment. In: SemEval. pp. 241--246.

\bibitem[{Turchi and Negri(2013)}]{turchi2013altn}
Turchi, M., Negri, M., 2013. Altn: Word alignment features for cross-lingual
  textual entailment. In: SemEval. pp. 128--132.

\bibitem[{Vuli\'{c} and Moens(2015)}]{DBLP:conf/sigir/VulicM15}
Vuli\'{c}, I., Moens, M., 2015. Monolingual and cross-lingual information
  retrieval models based on (bilingual) word embeddings. In: SIGIR. pp.
  363--372.

\bibitem[{Yih et~al.(2011)Yih, Toutanova, Platt, and Meek}]{yih:2011}
Yih, W., Toutanova, K., Platt, J.~C., Meek, C., 2011. Learning discriminative
  projections for text similarity measures. In: CoNLL. pp. 247--256.

\end{thebibliography}





\end{document}